\documentclass[preprint,12pt]{elsarticle}

\usepackage{amssymb}
\usepackage{amsmath}

\usepackage{amsmath,amssymb,amsfonts}
\usepackage{algorithmic}
\usepackage{graphicx}
\usepackage{textcomp}
\usepackage{xcolor}
\usepackage{algorithm}
\usepackage{gensymb}
\usepackage{booktabs}
\usepackage{verbatim}
\usepackage{subcaption}
\usepackage{caption}
\usepackage{comment}
\usepackage{makecell}

\journal{Robotics and Autonomous Systems}

\begin{document}

\begin{frontmatter}



\title{Hybrid Visual Servoing of Tendon-driven Continuum Robots}


\author[label1]{Rana Danesh} 
\author[label1]{Farrokh Janabi-Sharifi}
\author[label2]{Farhad Aghili}

\affiliation[label1]{organization={Department of Mechanical, Industrial, and Mechatronics Engineering, Toronto Metropolitan
University},
            addressline={ 350 Victoria Street}, 
            city={Toronto},
            postcode={M5B 2K3}, 
            state={Ontario},
            country={Canada}}

\affiliation[label2]{organization={Department of Mechanical, Industrial and Aerospace Engineering (MIAE), Concordia University},
            addressline={1455 Blvd. De Maisonneuve Ouest}, 
            city={Montreal},
            postcode={H3G 1M8}, 
            state={Quebec},
            country={Canada}}
            
\begin{abstract}
This paper introduces a novel Hybrid Visual Servoing (HVS) approach for controlling tendon-driven continuum robots (TDCRs). The HVS system combines Image-Based Visual Servoing (IBVS) with Deep Learning-Based Visual Servoing (DLBVS) to overcome the limitations of each method and improve overall performance. IBVS offers higher accuracy and faster convergence in feature-rich environments, while DLBVS enhances robustness against disturbances and offers a larger workspace. By enabling smooth transitions between IBVS and DLBVS, the proposed HVS ensures effective control in dynamic, unstructured environments. The effectiveness of this approach is validated through simulations and real-world experiments, demonstrating that HVS achieves reduced iteration time, faster convergence, lower final error, and smoother performance compared to DLBVS alone, while maintaining DLBVS's robustness in challenging conditions such as occlusions, lighting changes, actuator noise, and physical impacts.
\end{abstract}


\begin{highlights}
\item HVS integrates IBVS and DLBVS for robust TDCR control in dynamic environments.
\item HVS outperforms DLBVS in iteration time, error reduction, and control smoothness.
\item Experimental validation confirms HVS effectiveness under occlusions and noise.

\end{highlights}

\begin{keyword}
Hybrid Visual Servoing\sep  Tendon-Driven Continuum Robots\sep  Image-Based Visual Servoing\sep Deep Learning-Based Visual Servoing.



\end{keyword}

\end{frontmatter}




\section{INTRODUCTION}

Continuum robots (CRs) have gained popularity due to their unique flexible structure and adaptability, enabling them to operate effectively in unstructured environments \cite{1camarillo2008mechanics}. TDCRs are characterized by their small diameter-to-length ratios, making them ideal for navigating confined spaces \cite{2amanov2021tendon,3burgner2015continuum}.
The high degree of flexibility and numerous degrees of freedom of CRs present significant control challenges. Various control strategies, including both model-based \cite{4chikhaoui2018control} and model-free \cite{5george2018control} approaches, have been explored \cite{6da2020challenges}. The complexities involved in modeling and sensing further increase these control challenges. Developing accurate kinematic and dynamic models for CRs is an ongoing research challenge, often requiring iterative solutions to partial differential equations \cite{7till2019real,8janabi2021cosserat}. Additionally, sensing introduces its own set of challenges, such as size constraints, biocompatibility issues, and sterilization requirements \cite{9nazari2021image}. Consequently, non-contact sensing methods, particularly vision-based techniques, have become crucial in many CR applications \cite{10abdulhafiz2022deep}.

Vision-based control strategies offer attractive solutions by enabling effective sensing and direct endpoint manipulation of CRs, helping to circumvent challenges related to structural and calibration uncertainties \cite{11fallah2020depth,12hutchinson1996tutorial,13janabi2010comparison}. Early vision-based control methods, known as classical visual servoing, relied on projecting geometric features within images. Among these techniques, IBVS has been widely implemented, reducing error by directly minimizing the difference between the current and desired image features within the image plane. This approach is particularly effective in applications where reliable feature extraction and tracking are feasible. Numerous research papers have explored the use of IBVS in the control of TDCRs \cite{10abdulhafiz2022deep,norouzi2021constrained,norouzi2021switching,xu2021visual,wang2013visual}. Despite its effectiveness, IBVS is sensitive to occlusions and changes in lighting, and it requires precise feature extraction and tracking.

Direct visual servoing removes the need for explicit feature extraction and tracking by using the entire image as input. Recent advances in deep learning have enabled the development of direct visual servoing methods that use convolutional neural networks (CNNs) to learn a mapping from the image directly to the control commands. This approach can handle complex visual scenes and is robust to various disturbances, such as occlusions and changes in lighting. For instance, the related study by Bateux et al. involved training a CNN on images from diverse scenes, enabling real-time control of a rigid-link manipulator \cite{bateux2017visual, bateux2018training}. Similarly, Felton et al. developed a deep network using a Siamese network to predict the velocity of a camera on a robot tip, trained on the ImageNet dataset \cite{felton2021siame}. Abdulhafiz et al's study demonstrated the implementation of DLBVS specifically for a tendon-driven continuum robot, utilizing input images to directly control robot motion \cite{10abdulhafiz2022deep}. This method was validated through simulations in varied environmental conditions and real-world testing. The results showcased strong performance in normal, shadowed, and occluded scenarios, underscoring the effectiveness of DLBVS. A comparative analysis with IBVS showed that while DLBVS offers a larger workspace and greater robustness against environmental uncertainties, it requires more iterations to converge and exhibits higher error \cite{10abdulhafiz2022deep}.

Motivated by our comparative analysis between DLBVS and  IBVS, it's clear that the previous approaches have distinct advantages and limitations \cite{10abdulhafiz2022deep}. DLBVS effectively processes the entire image, enhancing adaptability to dynamic environments and disturbances. With a larger workspace than IBVS, it enables an extended operational range. However, it encounters challenges like slower convergence and higher final errors. In contrast, IBVS achieves high accuracy and faster convergence but relies heavily on the visibility and continuity of image features, which makes it more susceptible to occlusions and environmental changes. This comparison highlights the necessity for a hybrid visual servoing approach that combines the robust environmental adaptability and larger workspace of DLBVS with the accuracy and efficiency of IBVS.

In this paper, we contribute by proposing the first HVS approach that integrates the strengths of DLBVS and IBVS. By merging the robust environmental adaptability and full-image processing capabilities of DLBVS with the correctness and faster convergence of IBVS, we aim to develop a control strategy that can effectively guide TDCRs through complex and uncertain environments.

To provide an overview of the paper's organization, Section 2 details the methods used, including IBVS, DLBVS, and the proposed HVS approach. Section 3 presents the simulation and experimental results, showcasing the performance and robustness of the HVS system. Finally, Section 4 offers the conclusion and highlights the key findings of the study.

\section{Methods}
In the methods section, the techniques employed for HVS is presented, covering the traditional control method, IBVS, and the deep learning approach, DLBVS. Furthermore, we introduce the hybrid approach, which integrates both strategies to enhance overall performance.

\subsection{Image-Based Visual Servoing (IBVS)}

Classical IBVS aims to minimize the pixel error between the current and target features. In this approach, four distinct feature points were selected, with each point represented by the coordinate pairs $(u,v)$.

Given the four feature points, the classical image jacobian for each individual feature is defined as follows:

\begin{equation}
\mathbf{L_x} = \begin{bmatrix}
\frac{f}{z} & 0 & -\frac{u}{z} & -\frac{uv}{f} & \frac{f^2 + u^2}{f} & -v \\
0 & \frac{f}{z} & -\frac{v}{z} & \frac{f^2 + v^2}{f} & -\frac{uv}{f} & u
\end{bmatrix}
\end{equation}

\noindent where, \( f \) is the camera's focal length, and \( z \) is the image depth. After computing the jacobian matrices for the four features, the overall image jacobian, \( \mathbf{J_{img}} \) is constructed as:

\begin{equation}
\mathbf{J_{img}} = \begin{bmatrix}
\mathbf{L_{x1}} & \mathbf{L_{x2}} & \mathbf{L_{x3}} & \mathbf{L_{x4}}
\end{bmatrix}^T.
\end{equation}

 To approximate the Jacobian matrix of the TDCR, \(\mathbf{J_{robot}}\), a finite difference method was employed \cite{leibrandt2015line}. As the joint space variables represent tendon displacements, \(q_1\) and \(q_2\), their changes, \(\mathbf{\Delta q}\), were set to 0.1~mm, ensuring sub-millimeter accuracy. Subsequently, the resulting interaction matrix, denoted as \(\mathbf{L_e}\), was computed as:

\begin{equation}
\mathbf{L_e} = \mathbf{J_{img}} \mathbf{H} \mathbf{J_{robot}},
\end{equation}

\noindent where

\begin{equation}
\mathbf{H} = \begin{bmatrix}
\mathbf{R}_{3 \times 3} & \mathbf{0}_{3 \times 3} \\
\mathbf{0}_{3 \times 3} & \mathbf{R}_{3 \times 3}
\end{bmatrix},
\end{equation}

\noindent and \( \mathbf{R} \) is the rotation matrix from the base frame to the end-effector frame.

The classical IBVS control law is then defined as:

\begin{equation}
\mathbf{\Delta q} = -\lambda \mathbf{L_e}^+ (\mathbf{s} - \mathbf{s^*}),
\end{equation}

\noindent where \(\mathbf{\Delta q}\) represents the change in tendon displacements, \(\lambda\) is a gain factor, \(\mathbf{s}\) denotes the current feature vector, \(\mathbf{s^*}\) is the target feature vector, and  \( \mathbf{L_e}^+ \) represents the pseudo-inverse of \( \mathbf{L_e} \).

This control law effectively minimizes the pixel error between the current and desired features, ensuring accurate and efficient control of the TDCR's tip

\subsection{Deep Learning-Based Visual Servoing (DLBVS)}
DLBVS utilizes the advancements in deep learning to enhance the control of TDCRs. By employing CNNs, DLBVS directly maps visual inputs to control commands, bypassing the need for explicit feature extraction and tracking \cite{10abdulhafiz2022deep}. The following sections detail the control law, neural network integration, dataset collection, and the training and validation processes used in this study.

\subsubsection{Control Law}
\begin{figure}[H]
\centering
\includegraphics[width=0.9\textwidth]{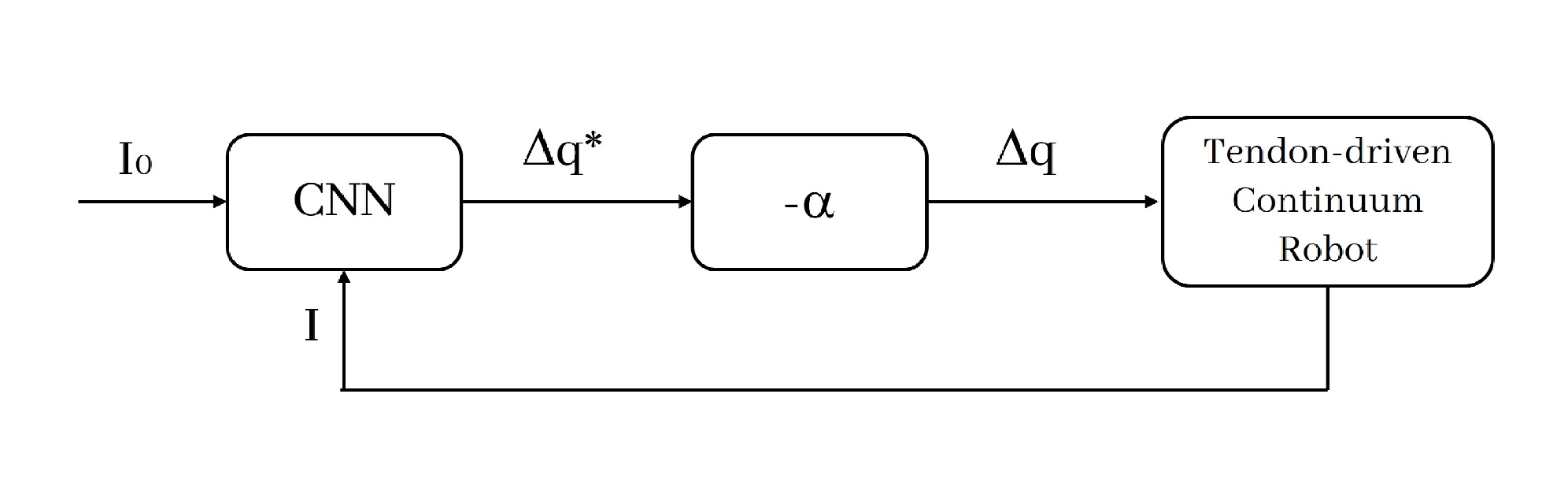}
\caption{Block diagram of the deep learning-based visual servoing controller.}
\label{fig4}
\end{figure}

The control strategy in DLBVS replaces the traditional mapping from image space to joint space with a trained neural network model. This model aims to minimize the error between the current image frame, \(\mathbf{I}\), and the target image frame, \(\mathbf{I_0}\). As depicted in Fig. \ref{fig4}, the model's output is scaled by a coefficient, \(-\alpha\), and then applied to the TDCR. The control law can be expressed with the following equations:

\begin{equation}
\mathbf{\Delta q^*} = f(\mathbf{I_0}, \mathbf{I})
\end{equation}

\begin{equation}
\mathbf{\Delta q} = -\alpha \mathbf{\Delta q^*}
\end{equation}

\noindent where, \(\mathbf{\Delta q}\) and \(\mathbf{\Delta q^*}\) represent the actual and desired changes in tendon displacements, respectively.

\subsubsection{Neural Network Integration}
A neural network was designed using a VGG-16 backbone pre-trained on the ImageNet dataset, utilizing transfer learning to optimize performance. This approach allows the network to utilize pre-learned features from natural images, requiring only the last layers of the network to be retrained to predict the required tendon displacements. In this model, the initial 10 layers were kept frozen to accelerate the training process. The final dense layer was removed and replaced with a new dense layer that outputs the two desired values corresponding to tendon displacements, $q_1$ and $q_2$, with a linear activation function applied to this layer \cite{10abdulhafiz2022deep}.

\subsubsection{Dataset Collection}

The dataset for training the neural network was generated using the Blender 3.6, which simulate the TDCR's environment and model the position and orientation of the end-effector (or camera) based on tendon displacements, $q_1$ and $q_2$. 

To ensure comprehensive coverage of the robot's workspace, a spiral path was used to traverse all reachable points in the 3D environment within a specified threshold. This path effectively stimulates the nonlinearities of the robot while covering all quadrants of the workspace. The farther the continuum robot is from the origin, the sparser the dataset becomes.

The spiral path was generated using the following equation:

\begin{equation}
q_1 = \frac{A}{n} x \cos \left( \frac{2\pi P}{n} x \right), 
q_2 = \frac{A}{n} x \sin \left( \frac{2\pi P}{n} x \right)
\end{equation}

\noindent where \( A \) is the maximum displacement of a tendon, \( P \) is the total number of periods the TDCR makes, \( n \) is the number of sample points, and \( x \) is an integer from 1 to \( n \). 

We introduced shadowing and occlusion effects to enhance the simulation's realism and robustness. Shadowing was managed by adjusting the light source, while occlusion was simulated by placing black rectangles randomly within the image. Using Blender's Python API, we automated the movement of the camera and light source, capturing numerous images of the scene from different angles and positions, thus creating a comprehensive dataset for training purposes.

 A total of 5000 images were acquired, each with a maximum amplitude of 10 mm and a period of 20. The input images were in RGB format and sized at 224×224 pixels.

\subsubsection{Training and Validation}

For the training process, the mean squared error (MSE) was selected as the loss function due to the linear activation function used for the output layer. This choice facilitated the learning of a direct mapping between the input images and the ground truth control points. The dataset's ground truth values were generated using a mapping that ensured they remained within the range of -1 to 1, which improved training efficiency and produced smoother convergence profiles. The mapping was defined as follows.

\begin{equation}
\mathbf{q_{mapped}} = \tanh(10\mathbf{q})
\end{equation}

The model was trained over 20 epochs, with a batch size of 32 and a learning rate of \(1 \times 10^{-4}\) using the Adam optimizer, achieving a final MSE of \(3.9 \times 10^{-4}\).

Training on a Core i7 CPU at 2.30 GHz with 16 GB of RAM took approximately 7.5 minutes per epoch, ensuring that the model effectively learned and produced accurate predictions for tendon displacements under varying conditions.

\subsection{Hybrid Visual Servoing (HVS)}

To quantitatively evaluate the error between the current and desired images, the pixel-wise Sum of Absolute Differences (SAD) is employed. This metric is computed between the normalized target and current images and is defined as follows:

\begin{equation}
\text{SAD} = \sum |\mathbf{I_0^*} - \mathbf{I^*}|
\end{equation}

\noindent where \(\mathbf{I^*}\) is the normalized current image and \(\mathbf{I_0^*}\) is the normalized target image. Here, the SAD value serves as a reliable metric for transitioning between IBVS and DLBVS.

The DLBVS approach provides an extended workspace and enhances the ability to manage TDCRs, even when there are significant differences between the current and target images. This suggests that IBVS functions effectively within a range where the SAD value remains below a certain threshold, while DLBVS maintains functionality across a broader spectrum of SAD values.

Another metric is the presence of features in the image. IBVS relies on a sufficient number of features to function effectively, and losing these features can lead to the failure of the controller. However, in situations where feature loss occurs, DLBVS can still guide the robot’s tip to the desired position.

Hybrid visual servoing operates based on two conditions: the SAD value at each iteration and the presence of features. If the SAD value falls below a predetermined threshold (selected based on the maximum SAD value at which IBVS can function), IBVS is activated; otherwise, DLBVS controls the robot. Additionally, if features become occluded, the system switches from IBVS to DLBVS. The hybrid visual servoing algorithm is presented in Algorithm \ref{algorithm 1}.

\begin{algorithm}[H]
\caption{Hybrid Visual Servoing}
\label{algorithm 1}
\begin{algorithmic}[1]
\STATE Initialize $\mathbf{I_0^*} \rightarrow \mathbf{q} = [0, 0]$
\STATE Initialize \( \mathbf{I^*} \rightarrow \mathbf{q} = \mathbf{q}_{\text{start}} \)
\STATE Calculate initial SAD

\FOR{iteration = 1 to N}
    \IF{SAD $<$ threshold (a)}
        \STATE \textbf{IBVS:}
        \STATE Calculate SAD and get features
        \IF{features are detected}
            \STATE Continue IBVS
        \ELSE
            \STATE Switch to DLBVS
        \ENDIF
    \ELSE
        \STATE \textbf{DLBVS:}
        \STATE Calculate SAD and get features
    \ENDIF
    \STATE Display \( \mathbf{q} \), \( \mathbf{\Delta q} \), SAD
\ENDFOR

\end{algorithmic}
\end{algorithm}

\section{Simulation and Experimental Results}
This section details the outcomes of the simulation and experimental studies designed to assess the performance of the HVS. It is structured to provide a comprehensive analysis, beginning with the experimental setup. The objective is to demonstrate the effectiveness of integrating DLBVS with IBVS in controlling TDCRs. The simulation studies provide valuable insights into the system's behavior, which are subsequently confirmed by the experimental results.

\subsection{Experimental Setup}
\begin{figure}[H]
\centering
\includegraphics[width=0.9\textwidth]{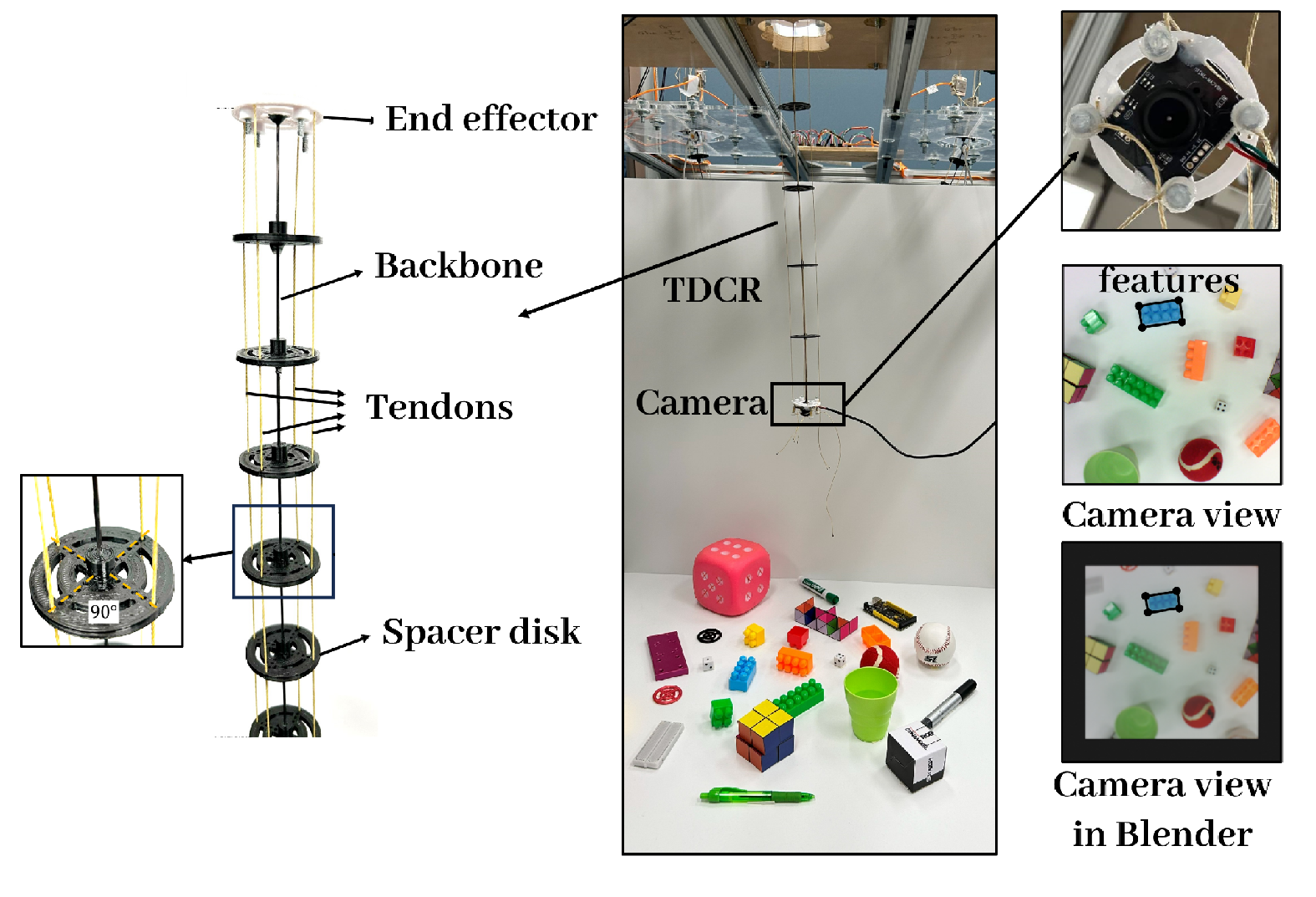}
\caption{The test setup consists of the TDCR with a camera attached to its tip.}

\label{fig1}
\end{figure}

The experimental setup, shown in Fig.~\ref{fig1}, consists of the TDCR with a camera attached to its end effector. The TDCR, kinematically modeled using the constant curvature assumption \cite{rao2021model}, is constructed with a 500 mm long spring steel backbone. This structure is equipped with four braided Kevlar lines (Emmakites, Hong Kong), each $0.45$ mm in diameter, serving as tendons. These tendons are spaced $20$ mm apart and arranged at $90$\textdegree$\ $intervals, guided by spacer disks fabricated from PLA filament. Actuation is achieved through XL430-W250-T servomotors (Dynamixel, CA).

The camera used in the setup is a USB camera module (Walfront, CN) featuring a $110$\textdegree$\ $wide-angle view and an OV3660 chip. It supports a resolution of $2048\times1536$, providing high-resolution, clear images with accurate color representation. The camera is attached to the tip using hot glue in an eye-in-hand (EIH) configuration.

For the calibration process, MATLAB is used to determine the camera parameters from chessboard images, providing the necessary focal length for IBVS and assuming a constant depth. Additionally, four blue block vertices in the scene are utilized as features in the IBVS method.

\subsection{Simulation}
\begin{figure}[H]
\centering
\includegraphics[width=0.99\textwidth, trim={5cm 9cm 3cm 7cm}, clip]{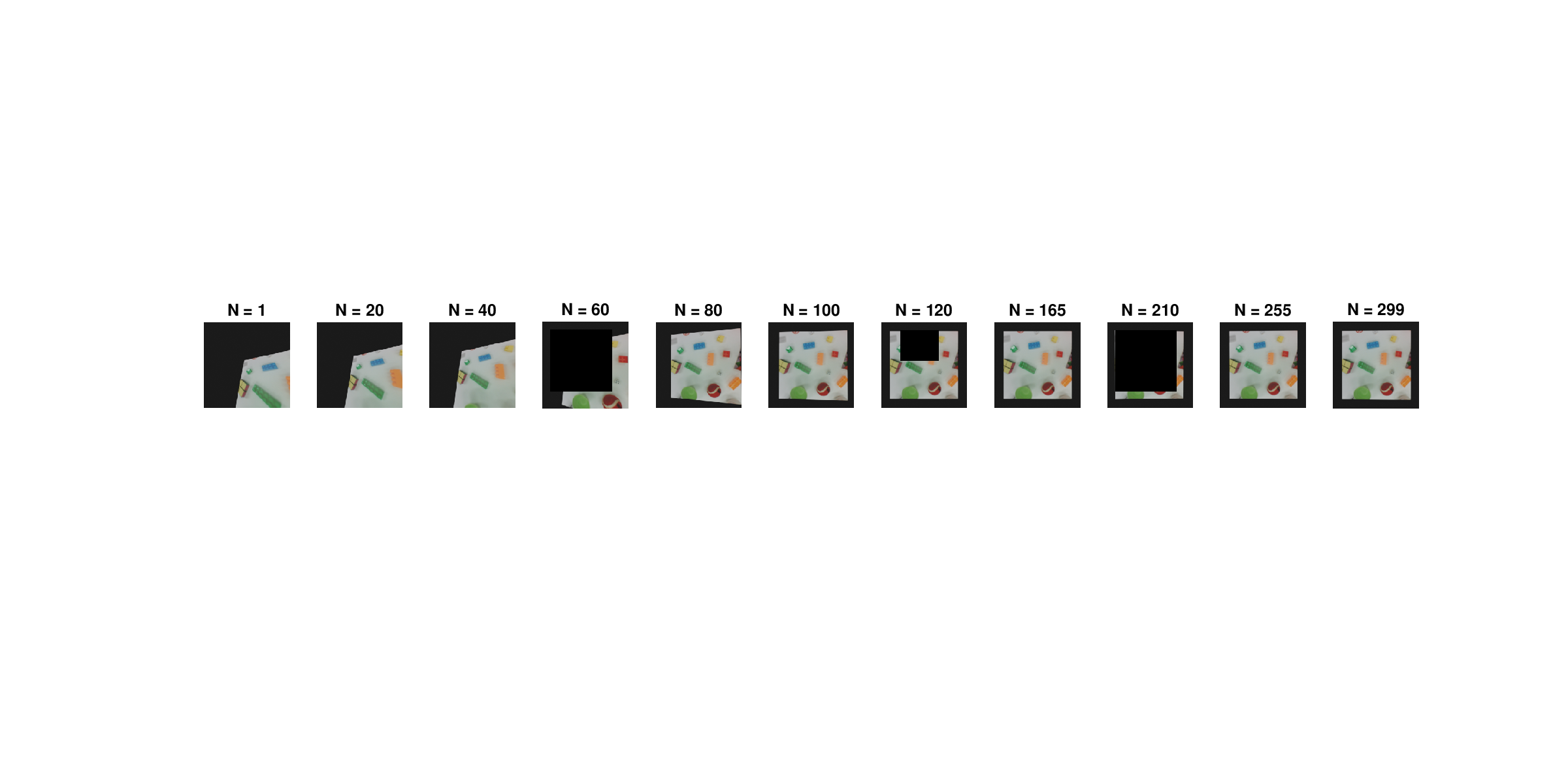}
\caption{The sequence of camera views in the simulation began with tendon displacements of \( (q_1, q_2) = (-10, 9)\) mm.
}
\label{Simulation_pics}
\end{figure}

Simulations were conducted in Blender software with initial tendon displacements of \( (q_1, q_2) = (-10, 9) \) mm, which are far from the target represented by \( (q_1, q_2) = (0, 0) \) mm. Fig. \ref{Simulation_pics} shows the initial image ($N=1$) that is the input to the HVS controller. Initially, the SAD value, calculated from the initial and target images, exceeded the threshold, causing the HVS to begin with DLBVS. To evaluate the efficiency and robustness of DLBVS, a single occlusion was applied to the images by placing a black rectangle of defined size and location between iterations 50 and 80 ($N=60$).

 After several iterations, as the SAD value dropped below the threshold, the system switched to IBVS. To observe the effect of missing features, additional occlusions were introduced between iterations 110 and 140, and 190 to 230 ($N=120$ and $210$), causing the HVS to revert to DLBVS. Once the occlusions were cleared, the system switched back to IBVS and continued until the end ($N=299$).

\begin{figure}[H]
\centering
\includegraphics[width=0.9\textwidth, trim=20 20 20 15, clip]{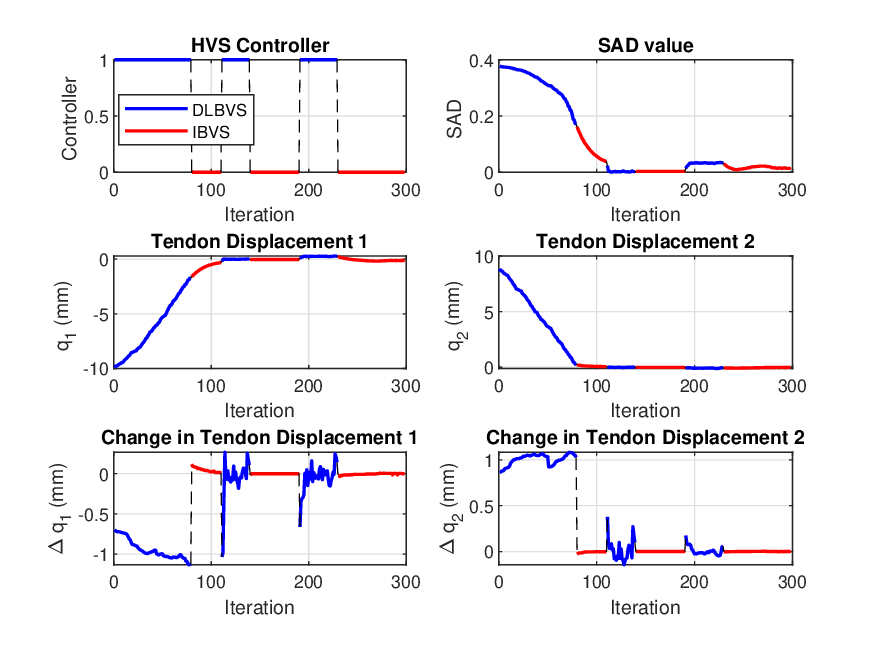}
\caption{The simulation results in Blender software demonstrate the performance of the HVS controller.}

\label{ALL_simulation}
\end{figure}

Fig.~\ref{ALL_simulation} illustrates the simulation results in Blender software. The top left figure shows the HVS controller in action, demonstrating how it initially started with DLBVS when the SAD value exceeded the threshold, and how it managed occlusions by switching between DLBVS and IBVS.

The top right figure presents the SAD value versus iteration, showing how the DLBVS initially controlled the TDCR and switched to IBVS once the SAD reached the defined threshold. Based on this figure, it can be concluded that IBVS exhibits less error and converges faster than DLBVS.

The middle left figure shows the displacement of \( q_1 \), which started at -10 mm and gradually approached 0, while the middle right figure shows second tendon displacement, \( q_2 \), which started from 9 mm. The IBVS segment demonstrates smoother behavior with less error, approaching zero more effectively than the DLBVS segment, although DLBVS is capable of handling occlusions.

This is further evident in the change in tendon displacements shown in the bottom figures. The oscillations are more prominent in DLBVS and exhibit greater error compared to IBVS. It should be noted that the DLBVS and the IBVS parts reflect the network output and control output, respectively.

\subsection{Experimental Validation}

In this section, the experimental validation of the HVS is presented. Its performance was evaluated through real-world tests using a TDCR, with a focus on the controller switching mechanism, SAD value, tendon displacements, and changes in tendon displacements. The validation is divided into three scenarios. Scenario 1 tests the HVS in real-world, consisting of four experiments conducted in different quadrants of the workspace in a normal condition. Scenario 2 evaluates the performance of HVS in comparison with DLBVS alone, as proposed in \cite{10abdulhafiz2022deep}. This scenario serves as a comparative analysis with an existing study, demonstrating the improvements achieved by HVS over prior DLBVS implementations. Scenario 3 assesses the robustness of HVS under various conditions, including occlusions, lighting changes, actuator noise, and physical disturbances.

\subsubsection{Scenario 1, HVS validation}
\begin{figure}[H]
\centering
\includegraphics[width=0.9\textwidth]{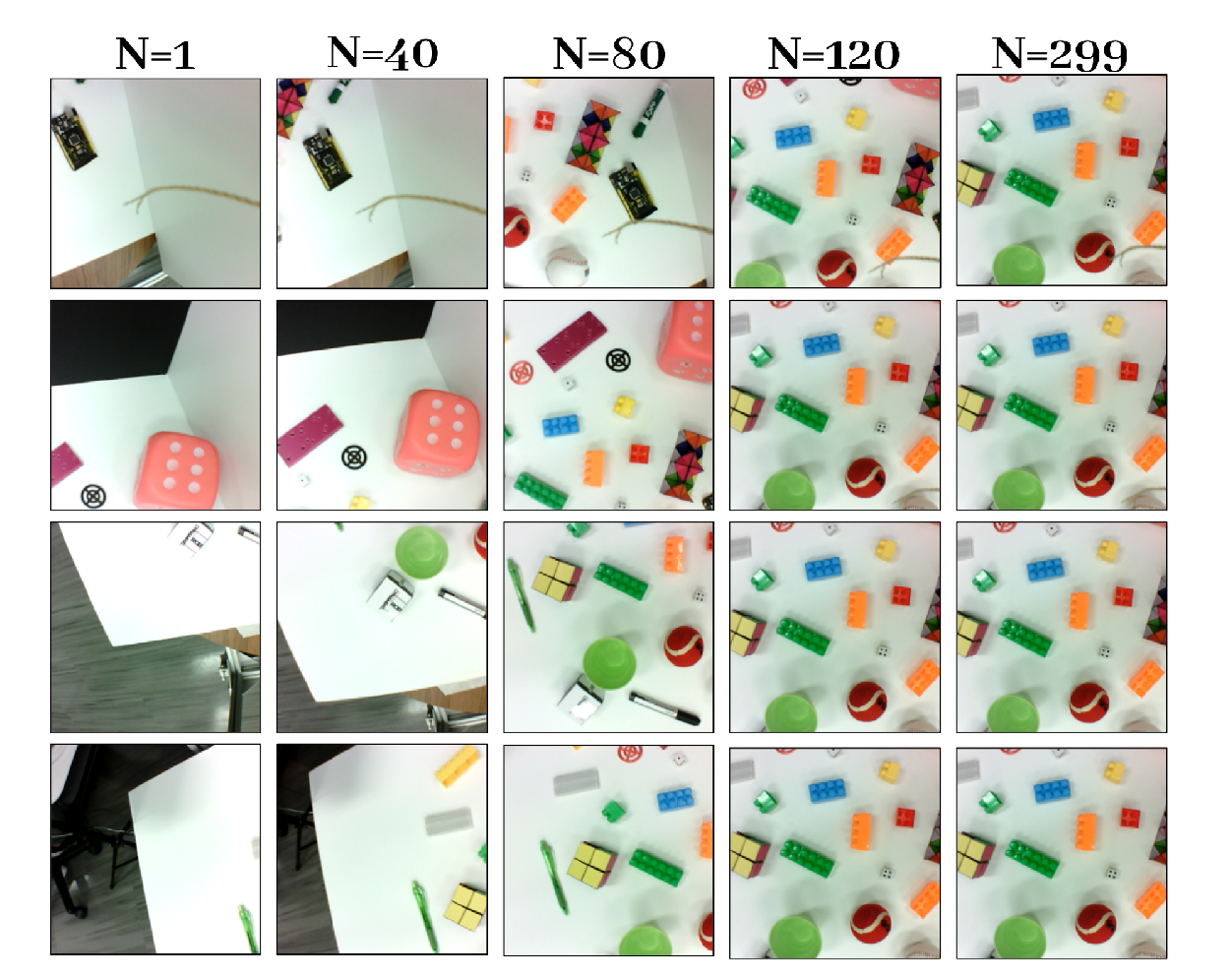}
\caption{Camera views under normal conditions with initial tendon displacements of \((10, 8)\), \((10, -8)\), \((-10, 8)\), and \((-10, -8)\).
}
\label{Senario1_images}

\end{figure}

Experiments were conducted with initial tendon displacements of \( (q_1, q_2) = (10, 8), (10, -8), (-10, 8), (-10, -8) \) mm, starting from positions far from the target, represented by \( (q_1, q_2) = (0, 0) \). The resulting camera views from the initial positions ($N=1$) to the final positions ($N=299$) are shown in Fig.~\ref{Senario1_images}.

\begin{figure}[H]
    \centering
    \begin{subfigure}[b]{0.49\textwidth}
        \centering
        \includegraphics[width=\textwidth, trim=40 0 33 0, clip]{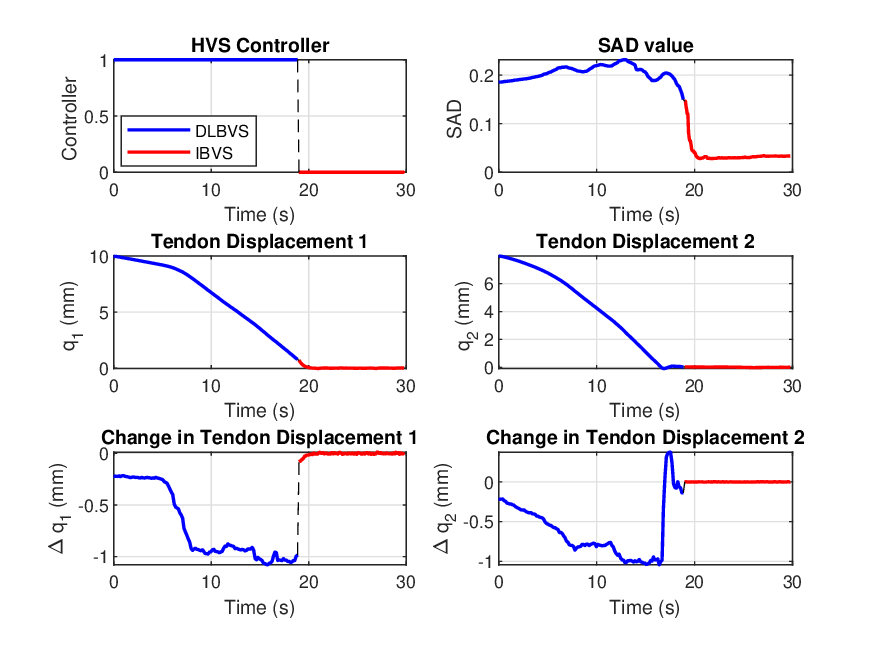}
        \caption{}
    \end{subfigure}
    \hfill
    \begin{subfigure}[b]{0.49\textwidth}
        \centering
        \includegraphics[width=\textwidth, trim=40 0 33 0, clip]{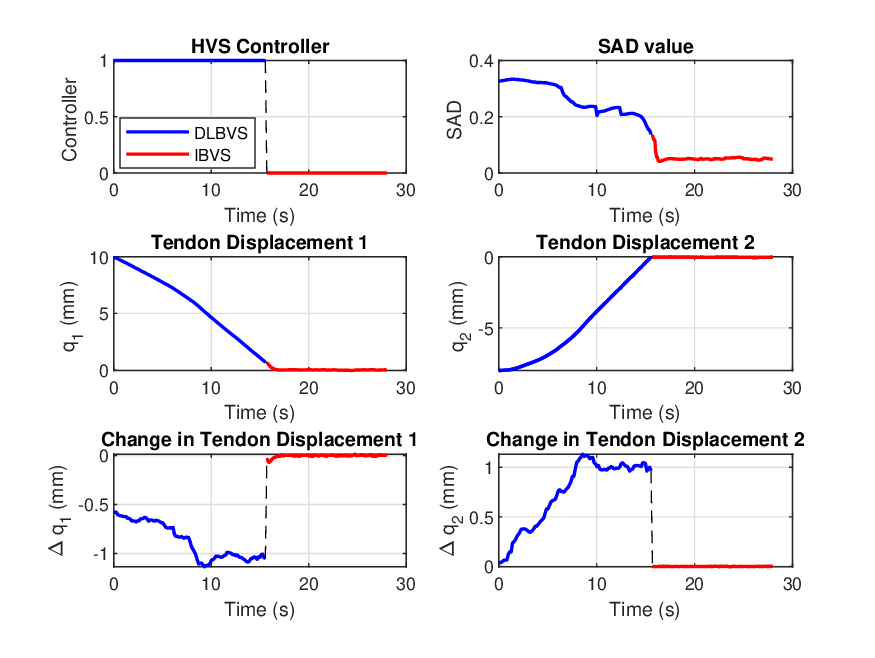}
        \caption{}
    \end{subfigure}
    \hfill
    \begin{subfigure}[b]{0.49\textwidth}
        \centering
        \includegraphics[width=\textwidth, trim=40 0 33 0, clip]{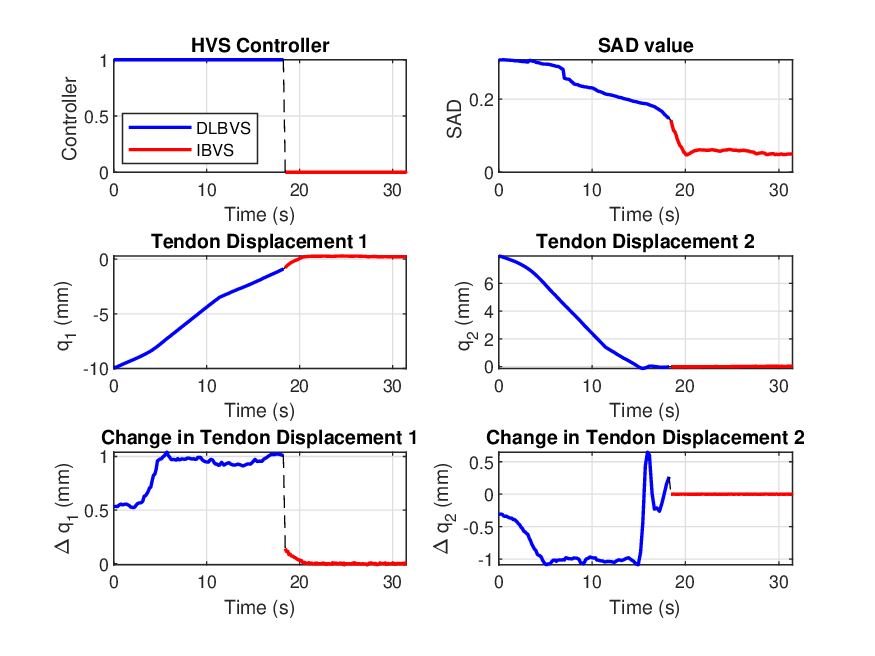}
        \caption{}
    \end{subfigure}
    \hfill
    \begin{subfigure}[b]{0.49\textwidth}
        \centering
        \includegraphics[width=\textwidth, trim=40 0 33 0, clip]{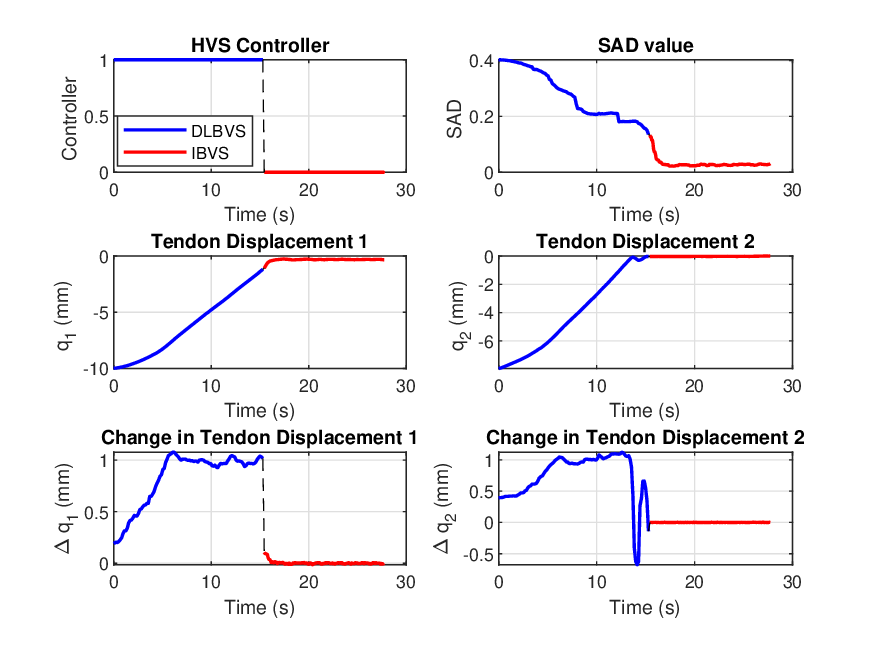}
        \caption{}
    \end{subfigure}

    \caption{The results of experiments conducted without any disturbances as \( (q_1, q_2) = \) (a) \( (10, 8)\), (b) \( (10, -8)\), (c) \( (-10, 8)\), and (d) \( (-10, -8)\).}
    \label{ALL_normal}
\end{figure}

All experiments were conducted without any disturbances, allowing the HVS to begin with DLBVS and switch to IBVS once the SAD value reached the defined threshold. The HVS controller, SAD value, tendon displacements, and their changes over time are shown in Fig.~\ref{ALL_normal}.

\subsubsection{Scenario 2, Comparison of HVS and DLBVS}
In this scenario, HVS and DLBVS begin the servoing process with the same initial tendon displacements of \( (q_1, q_2) = (10, -8) \). To quantitatively compare these two controllers, several metrics are considered: task completion, iteration time, convergence speed, final SAD value, and smoothness. Task completion determines whether the controller successfully drives the TDCR to the desired position. Iteration time measures the average duration (in seconds) of each iteration, with both controllers running for 300 iterations. A higher iteration time indicates greater computational demand and lower time efficiency. 

Convergence speed and final SAD are calculated from the SAD versus iteration graph, with a threshold of 0.06. The iteration at which the SAD value reaches and remains below 0.06 is recorded as the convergence speed, while the final SAD represents the SAD value at the last iteration. A lower final SAD indicates less final error. 

For smoothness, the standard deviation (std) and total path length (TPL) of change in tendon displacements ($[\Delta q_1, \Delta q_2]$) are compared. A higher standard deviation and total path length suggest less smooth control. 

Based on Table \ref{table2}, which compares HVS and DLBVS, both controllers successfully drive the TDCR to the desired position. However, HVS outperforms DLBVS across several metrics. The average iteration time for HVS is 0.0932 seconds, compared to 0.1443 seconds for DLBVS, demonstrating that integrating IBVS into DLBVS reduces the servoing time, making HVS more time-efficient. This is because DLBVS relies on CNNs, which require more processing time. 

Regarding convergence speed, HVS reaches the 0.06 threshold after 115 iterations, whereas DLBVS takes 175 iterations, indicating that IBVS accelerates the convergence in HVS. The final SAD value is 0.0493 for HVS and 0.0588 for DLBVS, showing that HVS achieves lower final error. In terms of smoothness, the standard deviation and total path length for both tendons are lower in HVS compared to DLBVS, indicating smoother control.

\begin{figure}[H]
    \centering
    \includegraphics[width=0.9\textwidth, trim=20 0 20 0, clip]{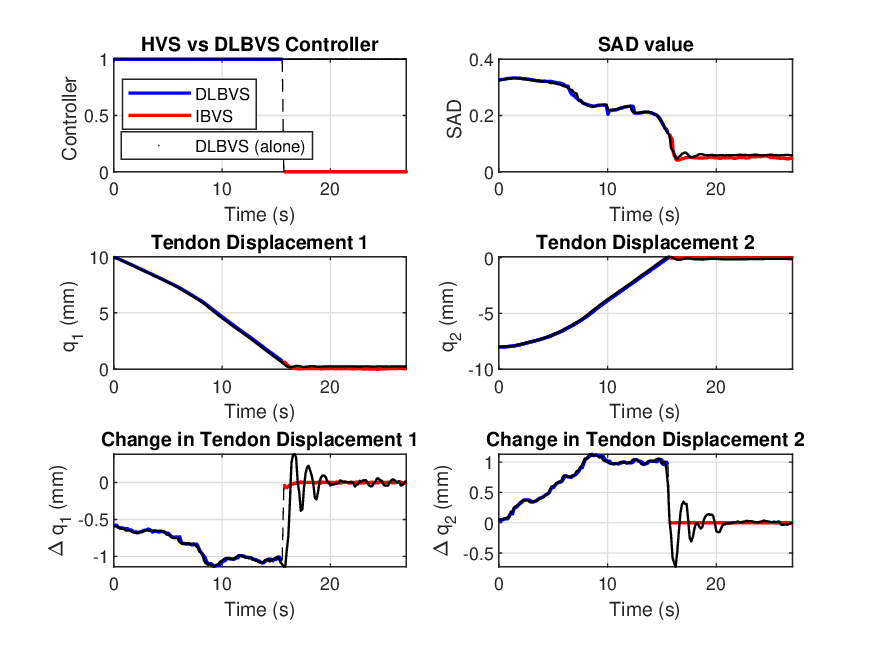}
    
    \caption{Comparison of HVS and DLBVS performance with initial tendon displacements of \( (q_1, q_2) = (10, -8)\).}
    \label{Compare_HVS}
\end{figure}

The performance comparison between the HVS and DLBVS is presented in Fig. \ref{Compare_HVS}. The results demonstrate the advantages of incorporating IBVS into the HVS controller, resulting in faster convergence, lower final error, and smoother operation, while maintaining the robustness and larger workspace benefits of DLBVS.

\begin{table}[H]
\centering
\caption{A Comparison between HVS and DLBVS.}
\resizebox{0.99\textwidth}{!}{%
\begin{tabular}{|c|c|c|c|c|c|}
\hline
\textbf{Controller} & \textbf{Task Completion} & \textbf{Iteration Time} & \textbf{Convergence Speed} & \textbf{Final SAD} & \textbf{Smoothness} \\ \hline
\textbf{HVS}        & Yes                     & 0.0932 s                       & 115 iterations             & 0.0493             & \makecell{std= [0.43107, 0.41013] \\ TPL= [2.9375, 3.2576]}  \\ \hline
\textbf{DLBVS}      & Yes                     & 0.1443 s                       & 175 iterations             & 0.0588             & \makecell{std= [0.44723, 0.42502] \\ TPL= [7.2822, 7.3266]}  \\ \hline
\end{tabular}%
}
\label{table2}
\end{table}

\subsubsection{Scenario 3, Robustness of HVS}
\begin{figure}[H]
\centering
\includegraphics[width=0.9\textwidth]{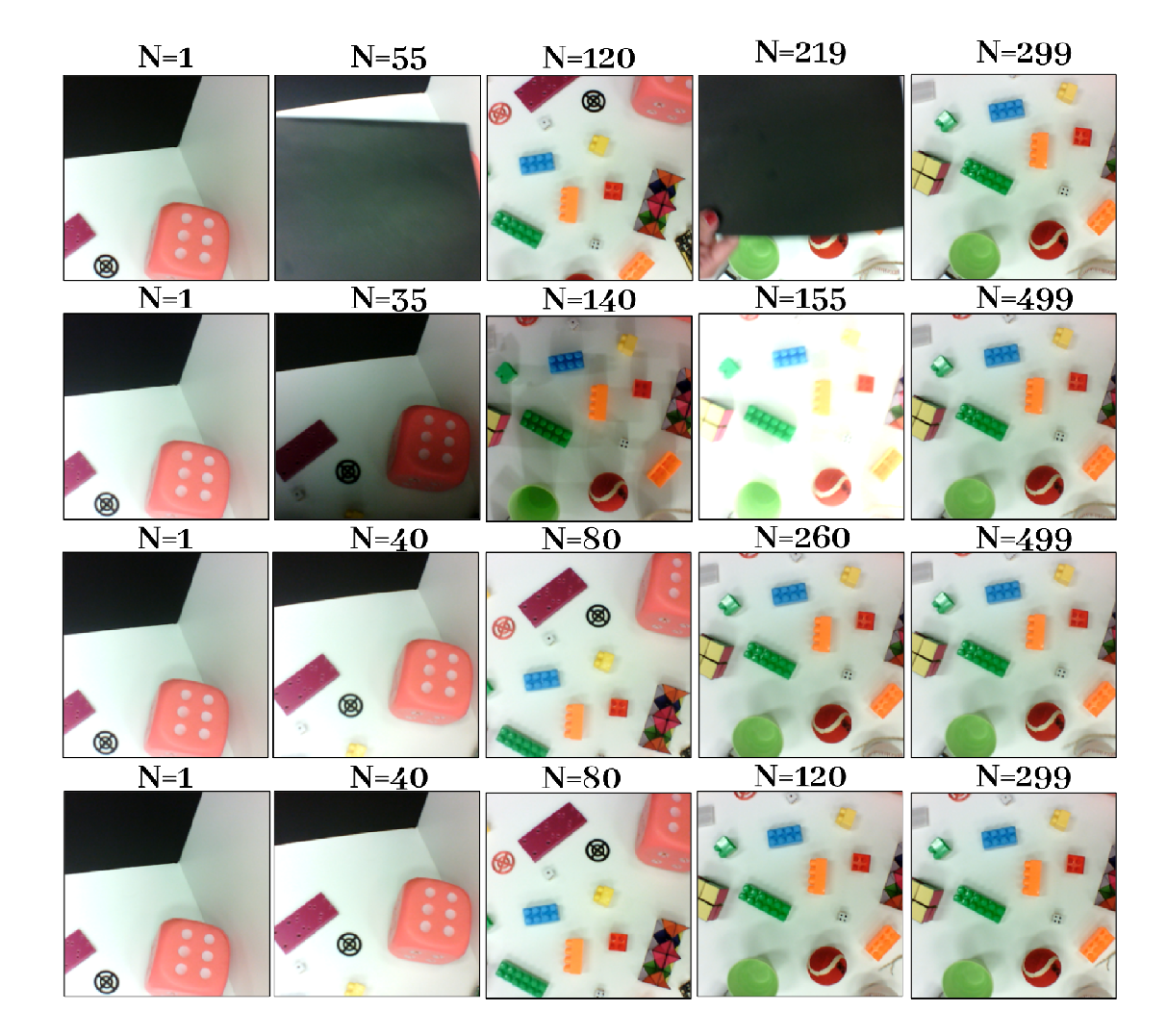}
\caption{Camera views from the HVS experiments designed to evaluate its performance under different types of disturbances. The rows show the effects of occlusion (the first row), lighting changes (the second row), actuator noise (the third row), and physical disturbance (the last row).}

\label{camera_robust}
\end{figure}

\begin{figure}[H]
    \centering
    \begin{subfigure}[b]{0.49\textwidth}
        \centering
        \includegraphics[width=\textwidth, trim=40 0 33 0, clip]{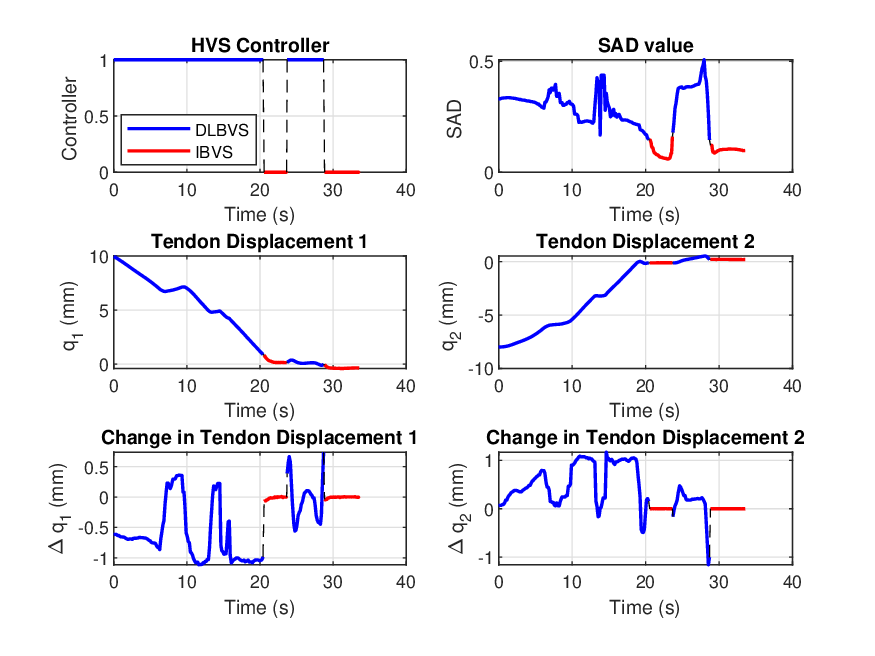}
        \caption{}
    \end{subfigure}
    \hfill
    \begin{subfigure}[b]{0.49\textwidth}
        \centering
        \includegraphics[width=\textwidth,trim=40 0 33 0, clip]{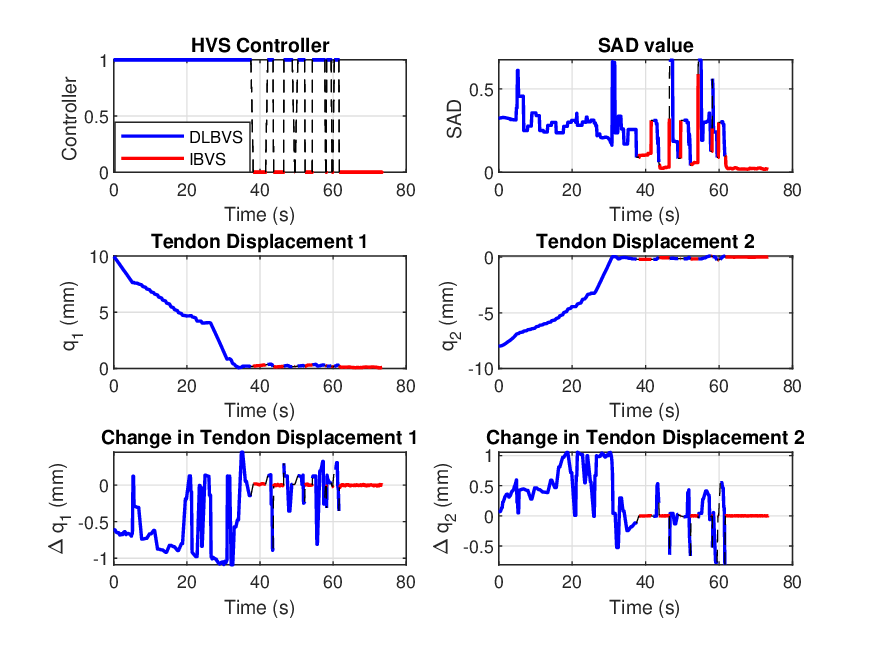}
        \caption{}
    \end{subfigure}
    \hfill
    \begin{subfigure}[b]{0.49\textwidth}
        \centering
        \includegraphics[width=\textwidth, trim=40 0 33 0, clip]{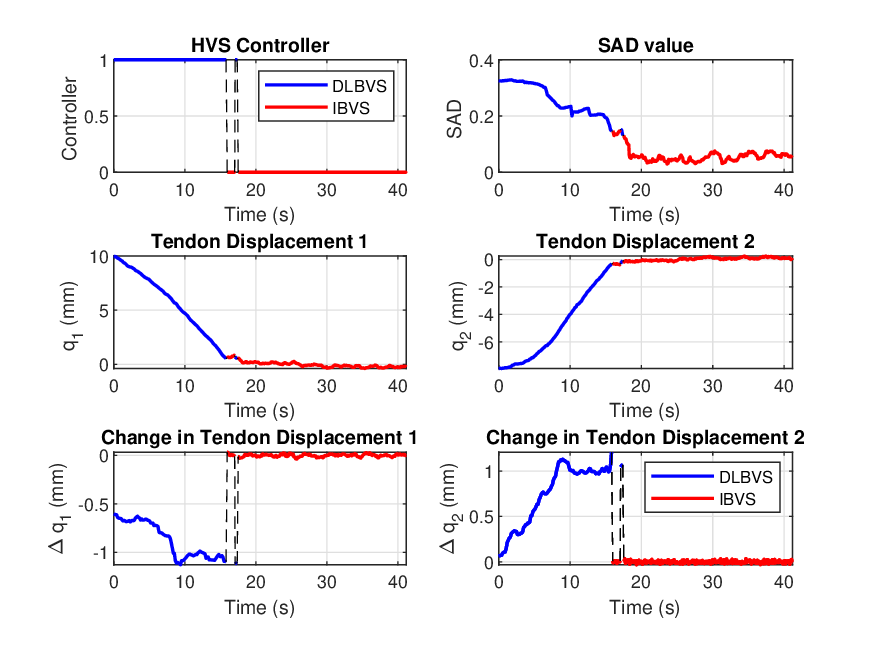}
        \caption{}
    \end{subfigure}
    \hfill
    \begin{subfigure}[b]{0.49\textwidth}
        \centering
        \includegraphics[width=\textwidth, trim=40 0 33 0, clip]{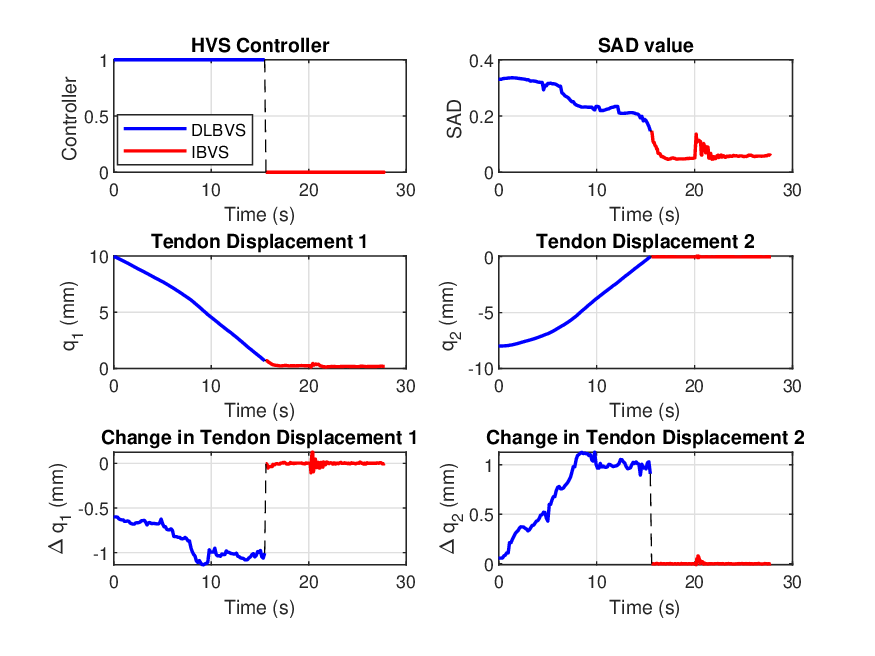}
        \caption{}
    \end{subfigure}

    \caption{Results of HVS experiments under various disturbances: (a) occlusions, (b) lighting changes, (c) actuator noise, and (d) physical disturbances.}
    \label{ALL_robust}
\end{figure}

In this scenario, four different disturbances were introduced. In the first experiment, occlusions were introduced, as shown in Fig.~\ref{camera_robust}, first row. When occlusion occurs during DLBVS, it continues without any sudden changes due to its robustness. However, during IBVS, if an occlusion happens, the features are lost, preventing IBVS from performing its task. At this point, it switches to DLBVS to handle the occlusion. Once the features become visible again, IBVS regains control of the TDCR, as shown in Fig.~\ref{ALL_robust}(a).

The second condition involved changes in lighting, achieved by randomly turning the lights on and off at varying frequencies. In Fig.~\ref{camera_robust}, the second row displays the camera view throughout the process, capturing normal lighting conditions along with shadowed and overexposed images. Fig.~\ref{ALL_robust}(b) illustrates how the HVS manages lighting changes by switching from IBVS to DLBVS when features cannot be detected, demonstrating the robustness of HVS in such situations.

The third condition involves actuator noise, introduced by adding white noise with a mean of zero and a standard deviation of 0.03 to the actuator signals, the tendon displacements \(q_1\) and \(q_2\). This condition is illustrated in the third row of Fig.~\ref{camera_robust} and Fig.~\ref{ALL_robust}(c).

The final disturbance involves applying physical impacts to the TDCR during the operation of the HVS. It is evident that the HVS remains robust under these conditions and continues the servoing process. Last row of Fig.~\ref{camera_robust} displays the camera views during the process, while Fig.~\ref{ALL_robust}(d) shows that, despite the physical impacts, the HVS successfully transitions from DLBVS to IBVS with minimal switching.

\section{Conclusion}

We proposed a novel HVS system that integrates IBVS and DLBVS to address the challenges of controlling TDCRs. The HVS approach leverages the strengths of both IBVS and DLBVS, enabling the system to handle unstructured environments with enhanced accuracy, faster convergence, greater robustness, and an expanded workspace. Through experiments, we demonstrated that HVS outperforms DLBVS alone in terms of iteration time, convergence speed, final error reduction, and smoothness of control, while retaining the robustness of DLBVS in challenging conditions such as occlusions, lighting changes, actuator noise, and physical disturbances.

\section{Acknowledgments}
This work was supported by the Natural Sciences and Engineering Research Council of Canada under Discovery Grants 2017-06930 and 2017-06764.




\begin{thebibliography}{00}


\bibitem{1camarillo2008mechanics}
  D. B. Camarillo, C. F. Milne, C. R. Carlson, M. R. Zinn, and J. K. Salisbury,
  \textit{Mechanics modeling of tendon-driven continuum manipulators},
  IEEE Transactions on Robotics, 
  Vol. 24, No. 6, pp. 1262--1273, 
  2008.


\bibitem{2amanov2021tendon}
  E. Amanov, T. Nguyen, and J. Burgner-Kahrs,
  \textit{Tendon-driven continuum robots with extensible sections—A model-based evaluation of path-following motions},
  The International Journal of Robotics Research, 
  Vol. 40, No. 1, pp. 7--23, 
  2021.

\bibitem{3burgner2015continuum}
  J. Burgner-Kahrs, D. C. Rucker, and H. Choset,
  \textit{Continuum robots for medical applications: A survey},
  IEEE Transactions on Robotics, 
  Vol. 31, No. 6, pp. 1261--1280, 
  2015.

\bibitem{4chikhaoui2018control}
  M. T. Chikhaoui and J. Burgner-Kahrs,
  \textit{Control of continuum robots for medical applications: State of the art},
  ACTUATOR 2018; 16th International Conference on New Actuators, 
  pp. 1--11, 
  2018.

\bibitem{5george2018control}
  T. G. Thuruthel, Y. Ansari, E. Falotico, and C. Laschi,
  \textit{Control strategies for soft robotic manipulators: A survey},
  Soft Robotics, 
  Vol. 5, No. 2, pp. 149--163, 
  2018.

\bibitem{6da2020challenges}
  T. Veiga, J. H. Chandler, P. Lloyd, G. Pittiglio, N. J. Wilkinson, A. K. Hoshiar, R. A. Harris, and P. Valdastri,
  \textit{Challenges of continuum robots in clinical context: a review},
  Progress in Biomedical Engineering, 
  Vol. 2, No. 3, pp. 032003, 
  2020.

\bibitem{7till2019real}
  J. Till, V. Aloi, and C. Rucker,
  \textit{Real-time dynamics of soft and continuum robots based on Cosserat rod models},
  The International Journal of Robotics Research, 
  Vol. 38, No. 6, pp. 723--746, 
  2019.

\bibitem{8janabi2021cosserat}
  F. Janabi-Sharifi, A. Jalali, and I. D. Walker,
  \textit{Cosserat rod-based dynamic modeling of tendon-driven continuum robots: A tutorial},
  IEEE Access, 
  Vol. 9, pp. 68703--68719, 
  2021.

\bibitem{9nazari2021image}
  A. A. Nazari, F.Janabi-Sharifi, and K. Zareinia,
  \textit{Image-based force estimation in medical applications: A review},
  IEEE Sensors Journal, 
  Vol. 21, No. 7, pp. 8805--8830, 
  2021.

\bibitem{10abdulhafiz2022deep}
  I. Abdulhafiz, A. A. Nazari, T. Abbasi-Hashemi, A. Jalali, K. Zareinia, S. Saeedi, and F. Janabi-Sharifi,
  \textit{Deep direct visual servoing of tendon-driven continuum robots},
  2022 IEEE 18th International Conference on Automation Science and Engineering (CASE), 
  pp. 1977--1984, 
  2022.

\bibitem{11fallah2020depth}
  M. M. H. Fallah, S. Norouzi-Ghazbi, A. Mehrkish, and F. Janabi-Sharifi,
  \textit{Depth-based visual predictive control of tendon-driven continuum robots},
  2020 IEEE/ASME International Conference on Advanced Intelligent Mechatronics (AIM), 
  pp. 488--494, 
  2020.

\bibitem{12hutchinson1996tutorial}
  S. Hutchinson, G. D. Hager, and P. I. Corke,
  \textit{A tutorial on visual servo control},
  IEEE Transactions on Robotics and Automation, 
  Vol. 12, No. 5, pp. 651--670, 
  1996.

\bibitem{13janabi2010comparison}
  F. Janabi-Sharifi, L. Deng, and W. J. Wilson,
  \textit{Comparison of basic visual servoing methods},
  IEEE/ASME Transactions on Mechatronics, 
  Vol. 16, No. 5, pp. 967--983, 
  2010.


\bibitem{norouzi2021constrained}
  S. Norouzi-Ghazbi, A. Mehrkish, M. M. H. Fallah, and F. Janabi-Sharifi,
  \textit{Constrained visual predictive control of tendon-driven continuum robots},
  Robotics and Autonomous Systems, 
  Vol. 145, pp. 103856, 
  2021.

\bibitem{norouzi2021switching}
  S. Norouzi-Ghazbi and F. Janabi-Sharifi,
  \textit{A switching image-based visual servoing method for cooperative continuum robots},
  Journal of Intelligent \& Robotic Systems, 
  Vol. 103, No. 3, pp. 42, 
  2021.

\bibitem{xu2021visual}
  F. Xu, H. Wang, W. Chen, and Y. Miao,
  \textit{Visual servoing of a cable-driven soft robot manipulator with shape feature},
  IEEE Robotics and Automation Letters, 
  Vol. 6, No. 3, pp. 4281--4288, 
  2021.
  
\bibitem{wang2013visual}
  H. Wang, W. Chen, X. Yu, T. Deng, X. Wang, and R. Pfeifer,
  \textit{Visual servo control of cable-driven soft robotic manipulator},
  2013 IEEE/RSJ International Conference on Intelligent Robots and Systems, 
  pp. 57--62, 
  2013.

\bibitem{bateux2017visual}
  Q. Bateux, E. Marchand, J. Leitner, F. Chaumette, and P. Corke,
  \textit{Visual servoing from deep neural networks},
  arXiv preprint arXiv:1705.08940, 
  2017.


\bibitem{bateux2018training}
  Q. Bateux, E. Marchand, J. Leitner, F. Chaumette, and P. Corke,
  \textit{Training deep neural networks for visual servoing},
  2018 IEEE International Conference on Robotics and Automation (ICRA), 
  pp. 3307--3314, 
  2018.

  
\bibitem{felton2021siame}
  S. Felton, E. Fromont, and E. Marchand,
  \textit{Siame-SE(3): Regression in SE(3) for end-to-end visual servoing},
  2021 IEEE International Conference on Robotics and Automation (ICRA), 
  pp. 14454--14460, 
  2021.

\bibitem{leibrandt2015line}
  K. Leibrandt, Ch. Bergeles, and G. Zh. Yang,
  \textit{On-line collision-free inverse kinematics with frictional active constraints for effective control of unstable concentric tube robots},
  2015 IEEE/RSJ International Conference on Intelligent Robots and Systems (IROS), 
  pp. 3797--3804, 
  2015.
  

\bibitem{rao2021model}
  P. Rao, Q. Peyron, S. Lilge, and J. Burgner-Kahrs,
  \textit{How to model tendon-driven continuum robots and benchmark modelling performance},
  Frontiers in Robotics and AI, 
  Vol. 7, pp. 630245, 
  2021.












\end{thebibliography}



\end{document}